\documentclass{bmvc2k}
\usepackage{graphicx}
\usepackage{comment}
\usepackage{wrapfig}
\usepackage{amsmath}
\usepackage{amssymb}
\usepackage{mathtools}
\usepackage{amsfonts}
\usepackage{booktabs}
\usepackage{caption}
\usepackage{multirow}
\usepackage{bm}
\usepackage{microtype}
\usepackage{subcaption}
\usepackage[rightcaption]{sidecap}

\usepackage{hyperref}
\hypersetup{
    colorlinks=true,
    linkcolor=blue,
    citecolor=blue,
    urlcolor=blue
}

\title{Unsupervised Landmark Discovery Using Consistency Guided Bottleneck}

\addauthor{Mamona Awan}{mamonaawan@yahoo.com}{1}
\addauthor{Muhammad Haris Khan}{muhammad.haris@mbzuai.ac.ae}{1}
\addauthor{Sanoojan Baliah}{sanoojan.baliah@mbzuai.ac.ae}{1}
\addauthor{Muhammad Ahmad Waseem}{msee20009@itu.edu.pk}{2}
\addauthor{Salman Khan}{salman.khan@mbzuai.ac.ae}{1,3}
\addauthor{Fahad Shahbaz Khan}{fahad.khan@mbzuai.ac.ae}{1,4}
\addauthor{Arif Mahmood}{arif.mahmood@itu.edu.pk}{2}

\addinstitution{
 Mohamed bin Zayed University of Artificial Intelligence (MBZUAI),\\
 UAE
}
\addinstitution{
 Information Technology University of the Punjab,\\
 Pakistan
}
\addinstitution{
Australian National University,\\
 Australia
}
\addinstitution{
 Link$\Ddot{\text{o}}$ping University,\\
 Sweden
}

\runninghead{M. Awan et al}{Unsupervised Landmark Discovery }

\begin{document}

\maketitle

\begin{abstract}

We study a challenging problem of unsupervised discovery of object landmarks. Many recent methods rely on  bottlenecks to generate 2D Gaussian heatmaps however, these are limited in generating informed heatmaps while training, presumably due to the lack of effective structural cues. Also, it is assumed that all predicted landmarks are semantically relevant despite having no ground truth supervision.
In the current work, we introduce a consistency-guided bottleneck in an image reconstruction-based pipeline that leverages landmark consistency~\textendash~a measure of compatibility score with the pseudo-ground truth~\textendash~to generate adaptive heatmaps. We propose obtaining pseudo-supervision via forming landmark correspondence across images. The consistency then modulates the uncertainty of the discovered landmarks in the generation of adaptive heatmaps which rank consistent landmarks above their noisy counterparts, providing effective structural information for improved robustness. Evaluations on five diverse datasets including MAFL, AFLW, LS3D, Cats, and Shoes demonstrate excellent performance of the proposed approach compared to the existing state-of-the-art methods. Our code is publicly available at {\url{https://github.com/MamonaAwan/CGB_ULD}}.
\end{abstract}

\section{Introduction}
\label{sec:intro}

Object landmark detection is an important computer vision problem. It portrays important information about the shape and spatial configuration of key semantic parts in 3D space for deformable objects like human and animal faces \cite{jakab2018unsupervised,sanchez2019object,mallis2020unsupervised,khan2020animalweb}. Many existing works have approached this problem in a fully-supervised manner \cite{bulat2017far,khan2017synergy,wang2019adaptive,miao2018direct,dong2018style,kumar2020luvli} which requires abundance of annotated images. Acquiring a large dataset of dense annotations for a particular object category may  be infeasible. Therefore, the current work aims discovering object landmarks in an unsupervised way.
Unsupervised learning of object landmarks is a challenging problem because the landmarks can express diverse configurations even for simple object categories like human faces. Also, recovering underlying mapping between spatial location and high-level semantic understanding of landmarks  without involving human supervision is quite challenging. Finally, the consistency of landmark detection should not be compromised under viewpoint variations, and detected landmarks should capture the shape of the deformable object \cite{sanchez2019object}.

\begin{figure*}[t]
\centering
        \begin{subfigure}[b]
        {0.48\textwidth}\label{fig:teaser_left}
        \includegraphics[width=1\linewidth]{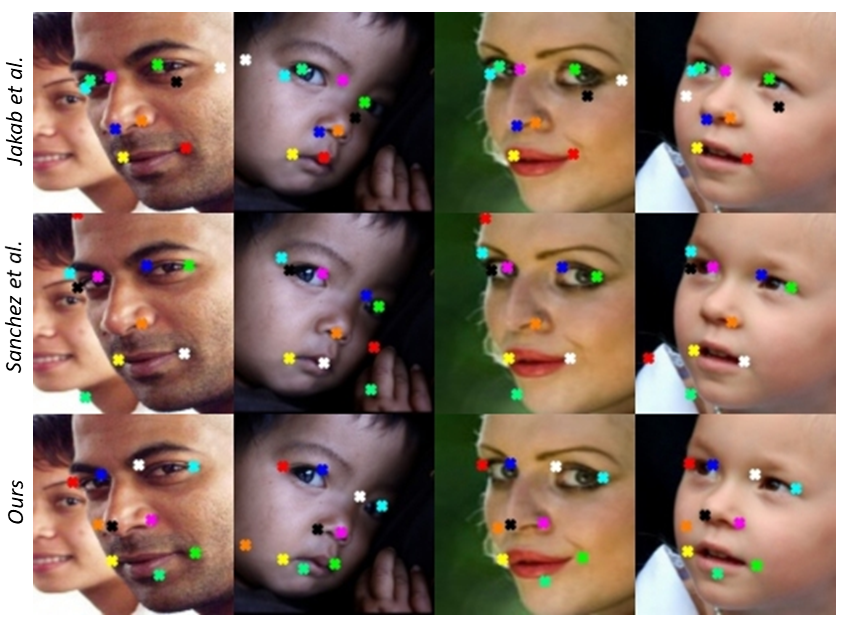}
        \end{subfigure}
        \hfill
        \begin{subfigure}[b]{0.48\textwidth}\label{fig:teaser_right}
        \includegraphics[width=1\linewidth]{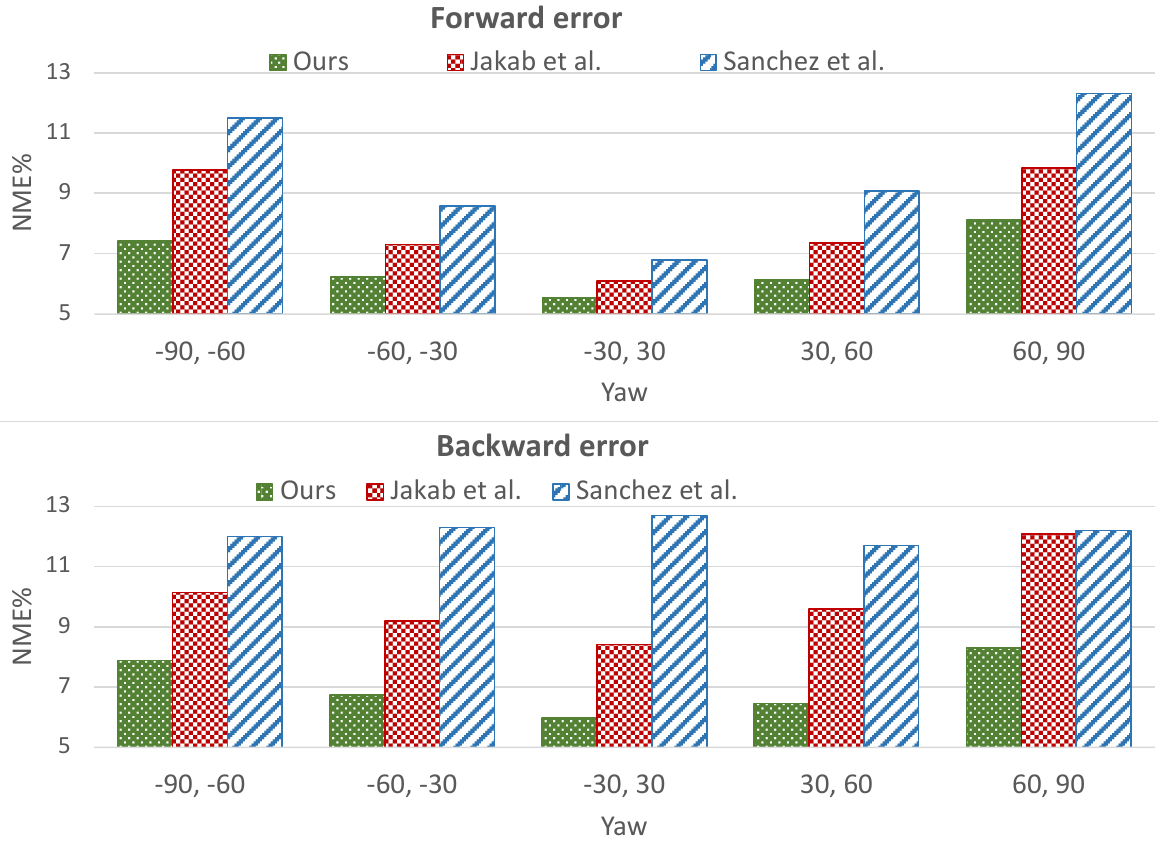}
        \end{subfigure}
      \caption{Left: Compared to ours, Jakab \textit{et al.}~\cite{jakab2018unsupervised} (top) and Sanchez \textit{et al.~}\cite{sanchez2019object} (middle) are prone to discovering semantically irrelevant landmarks lacking appropriate correspondence across varying poses. 
      Right: Comparison in terms of pose-wise NME(\%) based on yaw-angles on the AFLW\cite{koestinger2011annotated} dataset.}
         \label{fig:teaser_fig} 
\end{figure*}

Existing approaches to unsupervised landmark detection either impose equivariance constraint to 2D image transformation \cite{thewlis2017unsupervised,thewlis2019unsupervised,suwajanakorn2018discovery}, or leverage pre-text tasks such as (conditional) image generation \cite{jakab2018unsupervised,zhang2018unsupervised,sanchez2019object}. For instance, \cite{thewlis2017unsupervised} uses softargmax layer \cite{yi2016lift} to map the label heatmaps to a vector of points, and supervises the model with an equivariant error and a diversity constraint. Recently, Jakab \textit{et al.}~\cite{jakab2018unsupervised} proposed conditional image generation to guide learning of unsupervised landmark detection. They mapped the output of the softargmax layer to 2D Gaussian-like heatmaps using a \emph{bottleneck} which is tasked with distillation of object geometry, and hence it learns structured embeddings. These heatmaps are then utilized to reconstruct the input image from its deformed version. The bottleneck is a crucial component in their pipeline as it guides the landmark detector to detect landmarks which are able to effectively reconstruct a deformed version of the same image. Using the same pipeline, Sanchez \textit{et al.}~\cite{sanchez2019object} approached unsupervised landmark detection from a domain adaptation perspective via learning a projection matrix to adapt to new object categories. A problem inherent to these approaches  is that they cannot alleviate the impact of noisy structural cues, which can affect robustness under pose variations (see Fig.\ref{fig:teaser_fig}). We argue that a key reason is the naive formulation of the bottleneck. It assumes that, during training, all discovered landmarks by the detection network are equally meaningful under various variations. This is a  strict assumption, as it is likely that at least some discovered landmarks will be noisy. The resulting noisy structural cues can potentially limit the reconstruction ability and affect the robustness of landmark detector, making it detect semantically irrelevant landmarks, lacking appropriate correspondence (see Fig.\ref{fig:teaser_fig}).

In the current work, we address the aforementioned issues by introducing a \emph{consistency-guided bottleneck} formulation that leverages landmark consistency to generate adaptive heatmaps. We rank the discovered landmarks based on their consistency and hence favour relatively consistent ones. We obtain pseudo-supervision via establishing landmarks correspondence across the images. It includes clustering landmarks after estimating their confidence in a KNN affinity graph. This consistency is then used to modulate the uncertainty of the landmark in the generation of adaptive heatmaps. As a result, the adaptive heatmaps favour consistent landmarks over their counterparts, thereby providing effective structural cues while reconstructing the input image. This, in turn, facilitates the landmark detector to produce semantically meaningful landmarks. \footnote{\small Note that, the consistency-guided bottleneck facilitates detecting semantically meaningful landmarks and not semantic landmarks as such.} (see Fig.\ref{fig:teaser_fig}).

\noindent\textbf{Contributions:} \textbf{(1)} We introduce a novel \emph{consistency-guided bottleneck} formulation in the image reconstruction-based unsupervised landmark detection pipeline. It utilizes landmark consistency, a measure of affinity score with the pseudo-ground truth, for the generation of adaptive heatmaps. Such heatmaps potentially encode better structural information to facilitate an improved discovery of semantically meaningful and stable points. 
\textbf{(2)} We propose a principled way of generating adaptive heatmaps in an unsupervised mode. We first rank landmarks based on their consistencies and then modulate their corresponding uncertainties in the 2D Gaussian heatmaps. \textbf{(3)} We also introduce pseudo-supervision via establishing landmark correspondence across images. \textbf{(4)} Comprehensive experiments and analysis are performed on five  diverse datasets: MAFL, AFLW, LS3D, Cats, and Shoes. Our approach provides significant gains over the existing state-of-the-art methods.
\section{Related Work}
\label{sec:related_work}
\noindent \textbf{Unsupervised landmark detection methods}
can be broadly categorised into either imposing equivariance constraint to image transformations \cite{thewlis2019unsupervised,thewlis2017unsupervised_dense,thewlis2017unsupervised}, or leveraging image reconstruction as a pre-text task \cite{jakab2018unsupervised,sanchez2019object,jakab2020self}. In the absence of ground truth annotations, the equivariance constraint provides self-supervisory training signal. In particular, equivariance constraint requires representations across locations to be invariant to the geometric transformations of the image. 
Further constraints, based on locality \cite{thewlis2019unsupervised,thewlis2017unsupervised_dense} and diversity \cite{thewlis2017unsupervised} are introduced to avoid  trivial solutions. The generative methods  \cite{jakab2018unsupervised,lorenz2019unsupervised,sanchez2019object,wiles2018self,zhang2018unsupervised,bespalov2020brul,xu2020unsupervised,jakab2020self} employ equivariance constraints rather implicitly by considering objects as a deformation of the shape template in-tandem with the appearance variation in a disentangled manner \cite{cheng2021equivariant}. In \cite{zhang2018unsupervised}, landmark discovery is formulated as an intermediate step of image representation learning. Similarly, \cite{lorenz2019unsupervised} casts this as disentangling shape and appearance and introduced equivariance and invariance constraints into the generative framework. Wiles \textit{et al.}\cite{wiles2018self} proposed a self-supervised framework to embed facial attributes from videos and then utilized those to predict landmarks. Most of these methods observe lack of robustness under pose variations. 


\noindent \textbf{Deep clustering}  methods  employ clustering as pre-text task \cite{caron2018deep,li2016unsupervised,noroozi2018boosting,liao2016learning,yang2016joint} to partition the images into different clusters and a  classifier is trained to identify samples with same cluster id \cite{li2016unsupervised} or by using the cluster assignments as pseudo-labels \cite{noroozi2018boosting,caron2018deep}. For unsupervised landmark discovery, Mallis \textit{et al.}~\cite{mallis2020unsupervised} recovers landmark correspondence via k-means clustering and utilized them to select pseudo-labels for self-training in the first stage. The pseudo-labels are used to learn a landmark detector in a supervised manner in the second stage.
In contrast, we obtain pseudo-supervision to quantify landmark consistency. It is then used to modulate its 2D gaussian uncertainty in generating adaptive heatmaps. We do not use a dedicated feature head descriptor for learning landmark representations, and instead extract them directly from the encoder network. Moreover, we realize learning correspondence through clustering landmark representations after estimating their confidence in a KNN affinity graph. 


\begin{figure*}[!t]
    \centering    
    \includegraphics[width=\textwidth]{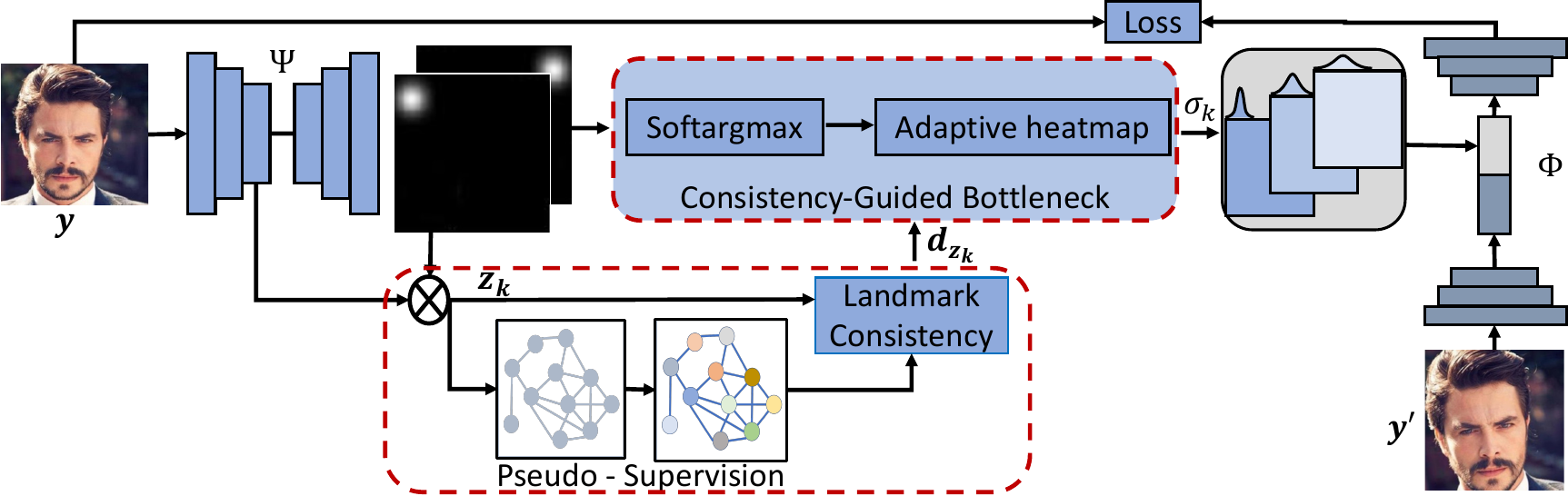}
    \caption{Overall architecture with consistency-guided bottleneck and pseudo-supervision.} 
    \label{fig:architecture}
    
\end{figure*}

\section{Proposed Consistency Guided Bottleneck}
\label{sec:overall_framework}
We aim to train a model capable of detecting landmarks for an arbitrary object category, without requiring ground truth annotations. 
Similar to the  prior works, we adopt an image generation based unsupervised landmark detection pipeline as shown in Fig.~\ref{fig:architecture}. It consists of a landmark detector network $\Psi$, and a generator network $\Phi$. 
An important part of this pipeline is conditional image generation  to guide the detection network in learning effective landmark representations. The object appearance in the first example image is combined with object landmark configuration in the second example image, where the two example images differ in viewpoint and/or object deformation.
Heatmap bottleneck is a crucial component in this pipeline for factorizing appearance and pose. It has a softargmax layer and a heatmap generation process. Specifically, the network $\Psi$ is terminated with a layer that ensures the output of $\Psi$ is a set of $k$ landmark detections. First, $k$ heatmaps are formed, one for each landmark, then each heatmap is renormalised to a probability distribution via spatial Softmax and condensed to a point by computing the spatial expected value. Finally, each heatmap is replaced with a Gaussian-like function centred at landmark location with a particular standard deviation depending upon the consistency of that landmark.
Although this unsupervised landmark detection pipeline shows encouraging results for some object categories, it struggles to detect
semantically meaningful landmarks, especially under large pose variations (Figs.~\ref{fig:teaser_fig},~\&~\ref{fig:LS3D_qual}). We believe the key reason is the naive formulation of the bottleneck, comprising of a softargmax layer and a heatmap generation process. The bottleneck assumes that all predicted landmarks 
are equally meaningful (i.e. have same semantic relevance). 
It is likely that at least some of the landmark detections will be noisy, particularly in the absence of ground truth supervision. To address this, we introduce a \emph{consistency-guided bottleneck} formulation that utilizes the landmark consistency towards generating adaptive heatmaps (Fig.~\ref{fig:architecture}).
\subsection{Consistency of a Landmark}
The consistency of a landmark is the proximity of its  representation to an assigned pseudo-label which is a cluster centroid in our case. As such, it allows us to rank landmarks based on their consistency measures and hence favour relatively consistent ones over inconsistent ones. We obtain pseudo-supervision via establishing correspondence of landmarks across images. The process includes clustering the landmark representations after estimating their respective confidences in a KNN affinity graph. The consistency is then used to modulate the uncertainty of the landmark's 2D gaussian to generate adaptive heatmaps. Consequently, the adaptive heatmaps allow reducing the impact of noisy structural information (e.g., unstable landmarks) while reconstructing the image, which in turn allows the landmark detector to produce semantically meaningful and stable landmarks. 
\subsection{Obtaining Pseudo-Supervision}
We obtain pseudo-supervision through establishing landmark correspondence across images. If two landmarks $k^i$ and $k^j$ in image $i$ and image $j$ correspond to the same semantic attribute (e.g. nose-tip), then their corresponding landmark representations $\mathbf{z}_{k}^{i}$, $\mathbf{z}_{k}^{j}$ should have the same pseudo-label. We realize this by clustering landmark representations after estimating their respective confidences in a KNN affinity graph.
We use the landmark representations to construct a KNN affinity graph $G=(V,E)$. Where each landmark representation is a vertex belonging to $V$, and is connected to its $\mathcal{K}$ nearest neighbors, forming $\mathcal{K}$ edges belonging to $E$. The affinity between landmark $k^i$ and landmark $k^j$ is denoted as $s_{i,j}$, which is the cosine similarity between their representations $\mathbf{z}_{k}^{i}$ and $\mathbf{z}_{k}^{j}$.

Using this affinity graph, we intend to perform the clustering of landmark representations by estimating the confidence of each landmark representation. The confidence reflects whether a landmark representation (a vertex in the affinity graph) belongs to a specific semantic attribute.
However, due to different variations in face appearance and pose, each landmark representation may have different confidence values even when they belong to the same semantic attribute (e.g., nose). 
For a landmark representation with high confidence, its neighboring landmark representations tend to belong to the same semantic attribute, while a landmark representation with low confidence is usually adjacent to the representations from the other landmarks.
Based on this, it is possible to obtain the confidence $c_{\mathbf{z}_{k}^{i}}$ for each landmark representation vertex based on the neighboring labeled representations as \cite{yang2020learning},
\begin{equation}
    c_{\bm{z}_{k}^{i}} = \frac{1}{|\mathcal{N}_{\bm{z}_{k}^{i}}|} \sum{}_{\bm{z}_{k}^{j}\in \mathcal{N}_{\bm{z}_{k}^{i}}} (\bm{1}_{y^j = y^i} - \bm{1}_{y^j \neq y^i}).s_{i,j},\end{equation}
where $\mathcal{N}_{\bm{z}_{k}^{i}}$ 
is the neighborhood of $\bm{z}_{k}^{i}$, $y^i$ is the ground truth label of $\bm{z}_{k}^{i}$ and $s_{i,j}$ is the affinity between $\mathbf{z}_{k}^{i}$ and $\mathbf{z}_{k}^{j}$. However, due to training in unsupervised mode, we cannot use aforementioned expression to compute the confidence for a landmark representation, and instead use a pre-trained graph convolutional network \cite{kipf2017semi} (GCN) to achieve the same.

With a pre-trained GCN, we can categorize the landmark representations based on their estimated confidences, to ultimately compute their cluster centroids. For a landmark representation vertex $\mathbf{z}_{k}^{i}$, neighbors with confidence larger than $\tilde{c}_{\mathbf{z}_{k}^{i}}$ show that they are more confident to belong to a certain cluster. Where $\tilde{c}_{\mathbf{z}_{k}^{i}}$ is the predicted confidence of $\mathbf{z}_{k}^{i}$. In this way, we assign each landmark representation to a cluster, and then compute the cluster-centroid by taking the mean of representations assigned to this cluster. We denote the number of cluster centroids by $T$ and they are much larger than the number of landmarks $K$ for capturing the intra-class variance in each semantic attribute \footnote{Note that, the value of $T$ is determined by the KNN+GCN clustering itself, and is set to 80 in Kmeans clustering.}. So, each semantic attribute could occupy more than one cluster.

\subsection{Quantifying landmark consistency}
We quantify the consistency of a landmark by relating it to each of the cluster centroids. In particular, given a landmark feature representation $\mathbf{z}_{k}$, we compute its similarity with the representations of $T$ cluster centroids and take the maximum similarity:
\begin{equation}
d_{\mathbf{z}_{k}}  = \mathrm{max}_{t \in T} \langle \mathbf{z}_{k}, \mathbf{z}_{t} \rangle,
\label{eq: d_k formula}
\end{equation}

\noindent where $\langle.,.\rangle$ is the cosine similarity operator, $\mathbf{z}_{t}$ is feature representation of $t^{th}$ cluster centroid, and $d_{\mathbf{z}_{k}}$ denotes the consistency of $k^{th}$ landmark.
We assume that, if a landmark representation $\mathbf{z}_{k}$ has higher similarity to its assigned cluster centroid, compared to another landmark representation, then it should be ranked higher in consistency compared to the other.
We empirically observed that our model's learning strives to improve landmark consistencies. Landmark consistency is also related to the performance, so the improvement in landmark consistency is corroborated by the decrease in error.
\subsection{Generating Adaptive Heatmaps}
We propose to generate adaptive 2D Gaussian heatmaps, as opposed to fixed ones, as it is likely that at least some proportion of the discovered landmarks will be noisy. In fixed heatmaps, the uncertainties of 2D Gaussians have a same constant value. This is particularly suitable if all landmark positions are semantically relevant, lying very close to the true spatial location of the semantic attribute. It is only possible if those landmarks are either carefully annotated by a human or perhaps, produced by some state-of-the-art fully-supervised landmark detector. However, in unsupervised mode, this is rather unlikely and hence we propose to rank these landmarks via modulating their 2D Gaussian uncertainties, to alleviate the impact of noisy landmarks in heatmap generation process.

Let $\Omega$ denote the image grid of size $H \times W$. The landmark detector $\Psi(\mathbf{y})$ produces $K$ heatmaps $S_u(\mathbf{y};k)$, $u \in \Omega$ one for each landmark $k = 1,...,K$. Where $u$ are the coordinates of a landmark.
These heatmaps are generated as the channels of a $\mathbb{R}^{H \times W \times K}$ tensor. We re-normalize each heatmap to a probability distribution using spatial softmax \cite{jakab2018unsupervised}: 
\begin{equation}
    u_{k}^{*}(\mathbf{y})=(\sum_{u\in \Omega}ue^{S_u(\mathbf{y};k)})/({\sum_{u\in \Omega}e^{S_u(\mathbf{y};k)}}).
\end{equation}
In this work, we allow each 2D gaussian in a heatmap to reflect landmark's consistency. In particular, we modulate the uncertainty $\sigma_{k}$ of 2D gaussian using the consistency $d_{\mathbf{z}_{k}}$ described in Eq. (\ref{eq: d_k formula}) as:$\sigma_k = 1 / exp(d_{\mathbf{z}_{k}})$.
Using this modulated uncertainty $\sigma_k$, we create \emph{adaptive heatmaps} by forming a Gaussian-like function, centred at the location of discovered landmark $k$ i.e. $u_k$.
\begin{equation}
\label{eq6}
\Psi_{u}(\mathbf{y};k) = \exp[-{1}/({2\sigma_k^2})||u-u_k^{*}(\mathbf{y})||^2]
\end{equation}
This results in a new set of $K$ adaptive heatmaps encoding the 2D Gaussian heatmaps the location of $K$ maximas, however, with a modulated uncertainty of 2D Gaussians reflecting landmark consistency. As such, this alleviates the impact of noisy landmark detections, thereby highlighting the consistent ones. These adaptive heatmaps then become input along with the deformed image representation to the reconstructor network $\Phi$. We observe that these adaptive heatmaps are a more informed encoding of spatial locations for the reconstructor network $\Phi$. This in turn  better facilitates the landmark detector $\Psi$ in producing semantically meaningful landmarks across poses and object categories. 


\section{Experiments}
\label{sec:experiments}

\noindent \textbf{Datasets:} We validate our approach on human faces, cat faces and shoes. For human faces, we use CelebA \cite{liu2015deep} (comprising of more than 200k celebrity images), AFLW \cite{koestinger2011annotated}, and the challenging LS3D \cite{bulat2017far} (containing large poses). For CelebA, we exclude the subset of test images of MAFL \cite{zhang2014facial}, which are used to test our trained models. For AFLW, we used the official train and test partitions. 
For LS3D, we follow the same protocol as in \cite{bulat2017far,mallis2020unsupervised} and use 300W-LP \cite{DBLP:journals/corr/ZhuLLSL15} for training. 
For cat faces, we choose Cats Head dataset \cite{zhang2008cat} ( 10k images). Following \cite{sanchez2019object}, we use 7,500 for training the landmark detector and the rest for testing. 
For Shoes, we choose UT-Zappos50k \cite{yu2014fine,yu2017semantic} (50k images), and use train/test splits from \cite{sanchez2019object}.   

\noindent \textbf{Landmark detector network:} We use the Hourglass architecture  \cite{newell2016stacked} as landmark detection network $\Psi$.
To obtain landmark representation, we concatenate the feature maps from the last block of encoder (768-D) and then reduce their dimensions to 256 using 1x1 convolution. 
The network produces heatmaps of spatial resolution $32 \times 32$, which are converted into $K\times 2$ tensor with a softargmax layer. We use element-wise multiplication of 256-D feature maps and heatmaps, to get 256-D representations of landmarks. For a fair comparison and following \cite{sanchez2019object}, the landmark detector $\Psi$ is initialised with the checkpoint, pre-trained on MPII.
For details on image reconstruction network, we refer to the supplementary material.



%
 
\noindent \textbf{Evaluation metrics:}
We use \emph{forward} error \cite{sanchez2019object,mallis2020unsupervised}, \emph{backward} error \cite{sanchez2019object}, and Normalised Mean-squared Error (NME), normalized by inter-ocular distance to report the performance. 


\begin{figure*}[b]
    \centering
\includegraphics[height=5cm,width=0.8\textwidth]{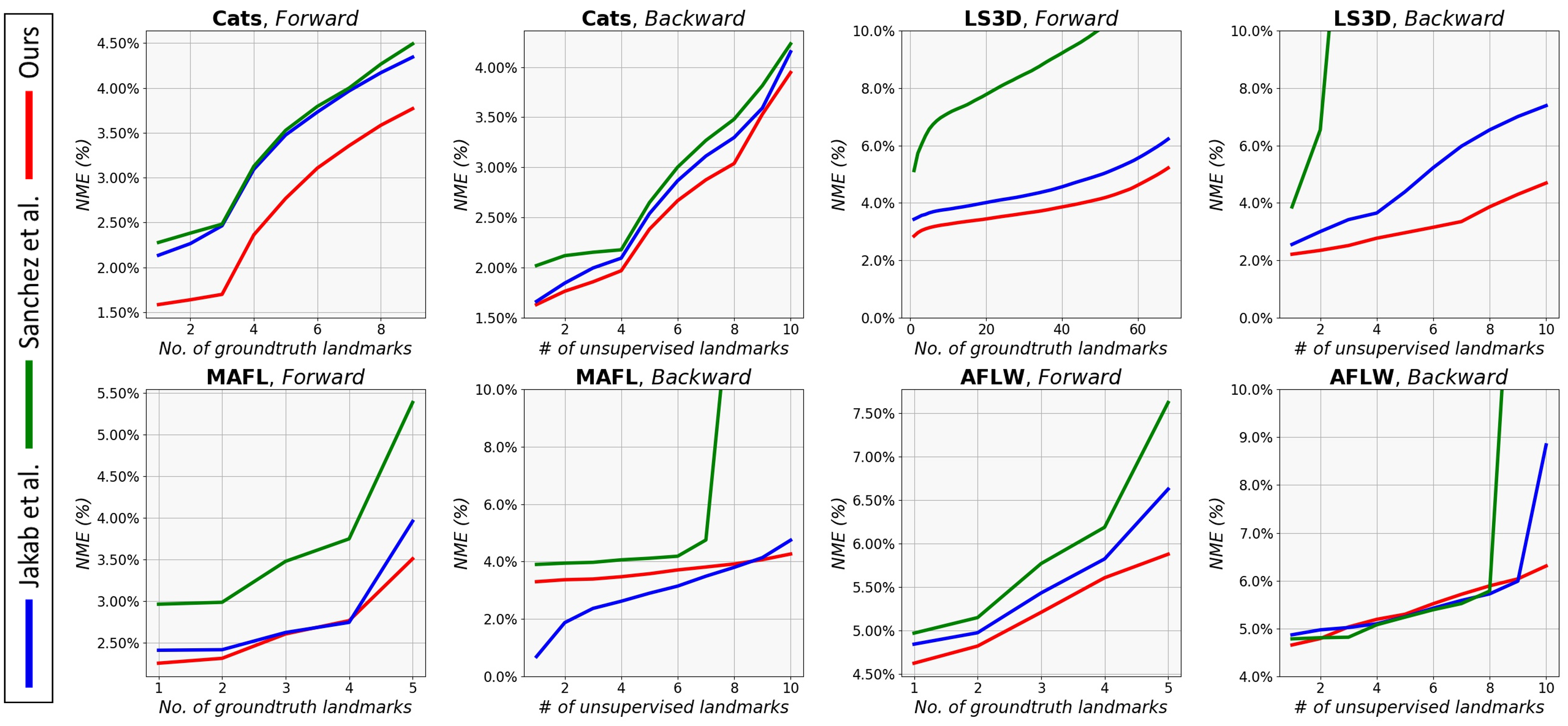}
    \caption{Cumulative error distribution (CED) curves for forward and backward errors.}
         \label{fig: Forwad_backward_error_comparison} 
\end{figure*}
\noindent \textbf{Training details:}
We use $\mathcal{K}=80$ in KNN affinity graph and use GCN to estimate confidences of the landmark representation vertices. In particular, we use a 1-layer pre-trained GCN on MS-Celeb-1M \cite{guo2016msceleb1m} dataset.
We obtain pseudo-supervision after every 5 epochs. Our overall network architecture is trained for 145 epochs, with a learning rate of $1\times 10^{-4}$, and a mini-batch size of 16 using Adam optimizer. 
\begin{SCtable}[\sidecaptionrelwidth][!htp]
    \tabcolsep=0.05cm
    {  \scalebox{0.65}{

\begin{tabular}{llcc}
\hline
\multicolumn{2}{l}{\text{Method}}                                           & \text{MAFL}                  & \text{AFLW}                  \\ \hline
\multicolumn{1}{l|}{\multirow{2}{*}{Sup}}          & TCDCN  \cite{zhang2015learning}                     & 7.95                           & 7.65                           \\
\multicolumn{1}{l|}{}                              & MTCNN  \cite{zhang2014facial}                       & 5.39                           & 6.90                           \\ \hline
\multicolumn{1}{c|}{\multirow{8}{*}{Unsupervised}} & Thewlis \cite{thewlis2017unsupervised}(K=30)        & 7.15                           & -                              \\
\multicolumn{1}{l|}{}                              & Jakab  \cite{jakab2018unsupervised}(K=10)$\dagger$  & 3.32                           & 6.99                           \\
\multicolumn{1}{l|}{}                              & Jakab  \cite{jakab2018unsupervised}(K=10)$\ddagger$ &  \textbf{3.19} & 6.86                           \\
\multicolumn{1}{l|}{}                              & Zhang  \cite{zhang2018unsupervised}                 & 3.46                           & 7.01                           \\
\multicolumn{1}{l|}{}                              & Shu  \cite{shu2018deforming}                        & 5.45                           & -                              \\
\multicolumn{1}{l|}{}                              & Shahasrabudhe  \cite{sahasrabudhe2019lifting}       & 6.01                           & -                              \\
\multicolumn{1}{l|}{}                              & Sanchez  \cite{sanchez2019object}                   & 3.99                           & 6.69                           \\
\multicolumn{1}{l|}{}                              & Mallis  \cite{mallis2020unsupervised}               & 4.12                           & 7.37                           \\ \hline
\multicolumn{1}{l|}{\multirow{2}{*}{Ours}}         & Baseline \cite{jakab2018unsupervised}               & 3.99                           & 7.03                           \\
\multicolumn{1}{l|}{}                              & Proposed                                                            & 3.50                           &  \textbf{5.91} \\ \cline{1-4} 
\end{tabular}} }
\caption{Performance comparison with the SOTA on MAFL and AFLW in forward errors. $\dagger$: uses the VGG-16 for perceptual loss, $\ddagger$: uses a pre-trained network for perceptual loss. Our method outperforms baseline by a notable margin in both datasets.} 
\label{tab:aflw_mafl_forward}
\end{SCtable}

\begin{figure*}
    \centering
    \includegraphics[width=0.95\linewidth]{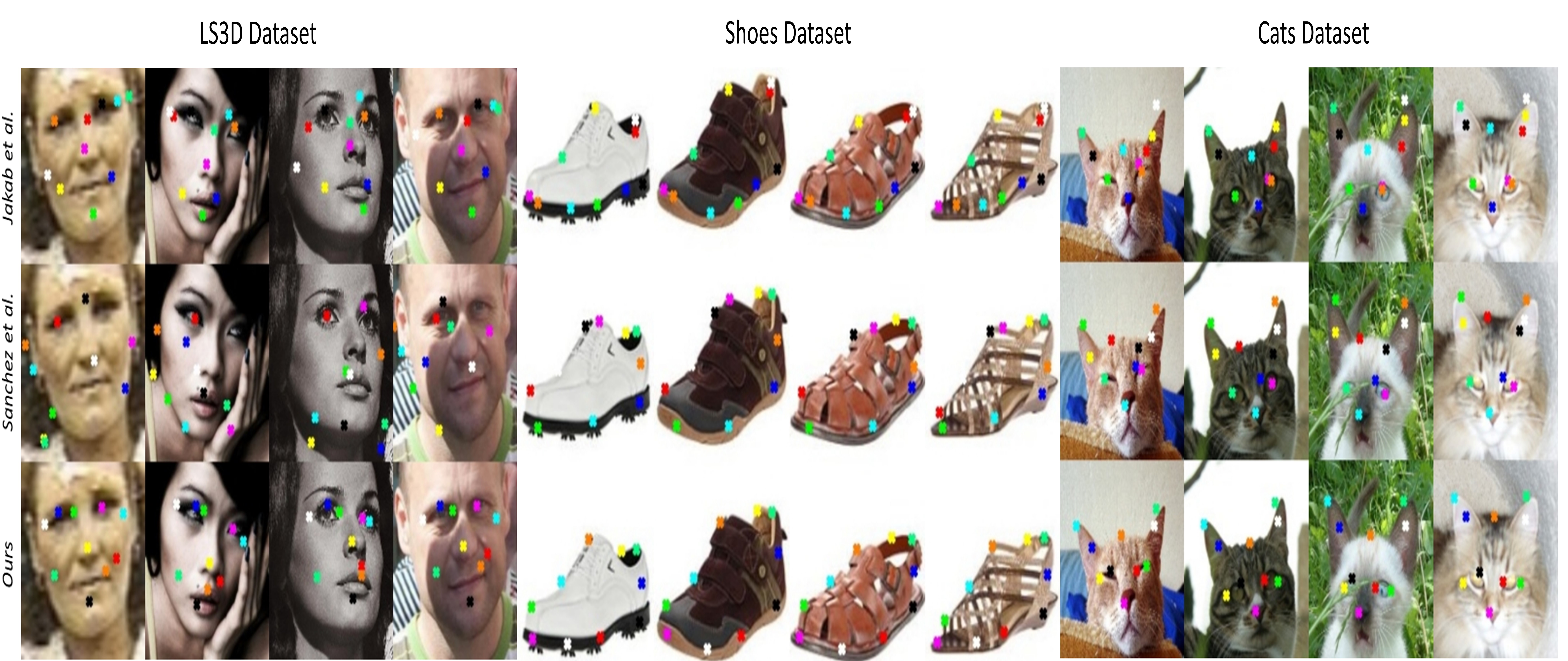}
    \caption{Visual comparison of ours with Jakab et al. \cite{jakab2018unsupervised} and Sanchez et al. \cite{sanchez2019object}. Our method discovers more semantically relevant landmarks and recovers improved correspondence.}
    \label{fig:LS3D_qual}
\end{figure*}

\noindent\textbf{{Comparison with the state-of-the-art (SOTA):\\}
}
\noindent\textbf{MAFL and AFLW:}  
In the forward error evaluation (Tab.~\ref{tab:aflw_mafl_forward}), our method outperforms the baseline by a notable margin in both MAFL and AFLW datasets. Furthermore, it provides a significant improvement over the recent top performing methods of \cite{sanchez2019object} and \cite{mallis2020unsupervised} in both datasets.  
Our baseline is an in-house implementation of the existing pipeline. 
In backward error evaluation (Tab.~\ref{tab:backward_aflw_mafl}), our approach demonstrates the best performance by achieving the lowest NME of 4.26\% and 6.39\% on MAFL and AFLW, respectively. See Fig.~\ref{fig: Forwad_backward_error_comparison} for Cumulative Error Distribution (CED) curves. \noindent\textbf{LS3D, Cats and Shoes:} In LS3D, our method achieves the best performance in both forward and backward errors (Tab.~\ref{Backward error forward error ls3d cats} (left)), and detects semantically meaningful landmarks with improved correspondence (Fig.~\ref{fig:LS3D_qual}). 
On Cats Head, our method delivers improved performance compared to others in both forward and backward errors (Tab.~\ref{Backward error forward error ls3d cats} (right)), and despite variations (e.g., appearance and expressions) it discovers landmarks displaying improved correspondence across images (Fig.~\ref{fig:LS3D_qual}).
\begin{SCtable}
    \centering
    \tabcolsep=0.05cm
    \small
    \scalebox{0.90}{
\begin{subtable}[c]{0.40\linewidth}
\centering
\tabcolsep=0.07cm
 {  \scalebox{1}{ \begin{tabular}{l c c c}
    \toprule
         Method & & Forw. Err. & Backw. Err.\\
    \toprule
        Baseline\cite{jakab2018unsupervised}                  &  & 5.38 & 7.06  \\
        Sanchez\cite{sanchez2019object}                      &  & 26.41  & 5.44  \\
        Mallis\cite{mallis2020unsupervised}                  &  & 6.53 & 6.57  \\
        Ours                                                  &  & \textbf{5.21} & \textbf{4.69}  \\
        \bottomrule
    \end{tabular}}

  }
   
\end{subtable}
\hfill
\begin{subtable}[c]{0.40\linewidth}
\centering
\tabcolsep=0.07cm
 {\scalebox{1}{\begin{tabular}{l c c c}
    \toprule
         Method & &Forw. Err. & Backw. Err.\\
    \toprule
        Baseline\cite{jakab2018unsupervised}                 &    & 4.53 & 4.06  \\
        Sanchez\cite{sanchez2019object}                    &    & 4.42  & 4.17  \\
        Ours                                                &    & \textbf{3.76} & \textbf{3.94}  \\
    \bottomrule
    \end{tabular}}}
\end{subtable}}
\caption{Error comparison  on (left) LS3D, (right) Cats Head  datasets.}
\label{Backward error forward error ls3d cats}
\end{SCtable}

\begin{SCtable}
    \centering
    \tabcolsep=0.05cm
    \small
    \centering
     \scalebox{0.9}{ 
    \begin{tabular}{l c c c}
    \toprule
         Method & &MAFL & AFLW\\
    \toprule
        Baseline\cite{jakab2018unsupervised}               &   & 4.53 & 8.84 \\
        Sanchez\cite{sanchez2019object}                      &   & 14.74  & 25.85\\
        Mallis\cite{mallis2020unsupervised}                  &   & 8.23  & - \\
        Ours                                                 &   & \textbf{4.26}  & \textbf{6.39}\\
    \bottomrule
    \end{tabular}}
    \caption{Backward errors comparison on MAFL and AFLW datasets.} 
\label{tab:backward_aflw_mafl}
\end{SCtable}

\noindent\textbf{Stability Analysis:} 
The stability of discovered landmarks is evaluated by measuring the error per landmark \cite{sanchez2019object} as, $e_{k} = ||\Psi_{k}(A(\mathbf{y})) - A(\Psi_{k}(\mathbf{y}))||$, where $A$ denotes a random similarity transformation. We report stability error, averaged over K=10 landmarks, in Tab.~\ref{tab:stability_errors}. Our method produces more stable landmarks than the competing approaches on most datasets.

\begin{SCtable}
    \centering
    \tabcolsep=0.05cm
    \small
    \scalebox{0.8}{
    \begin{tabular}{l c c c c c}
    \toprule
         Method & MAFL  & AFLW & Cats Head & LS3D & Shoes\\
    \toprule
        Baseline\cite{jakab2018unsupervised}                &\textbf{2.16}   & 3.12 &2.59 & 4.95  & 2.83 \\
        Sanchez\cite{sanchez2019object}  & 8.78   & 7.56  & 2.58 & 21.3 & 2.45 \\
        Ours &2.37 & \textbf{1.77} & \textbf{2.24} & \textbf{3.23} & \textbf{2.19}\\
    \bottomrule
    \end{tabular}}
    \caption{Stability errors for our method and the other two SOTA approaches.}
    \label{tab:stability_errors} 
\end{SCtable}

\begin{figure*}[t]

\centering
    \begin{subfigure}[Baseline]{0.32\textwidth}
         \centering
        \includegraphics[width=0.75\textwidth]{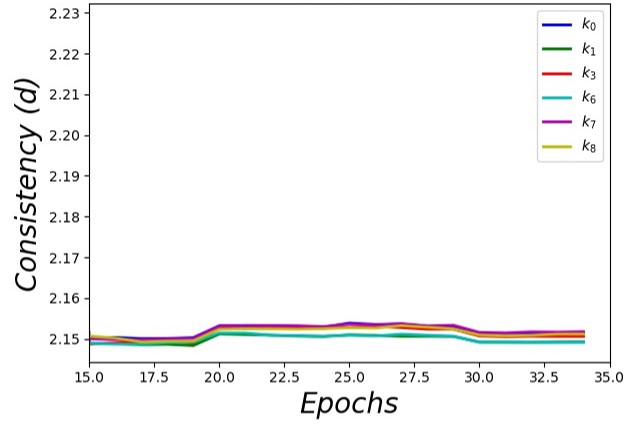}
        \label{fig:consistency_baseline}
     \end{subfigure}
     \begin{subfigure}[Ours]{0.32\textwidth}
         \centering
        \includegraphics[width=0.75\textwidth]{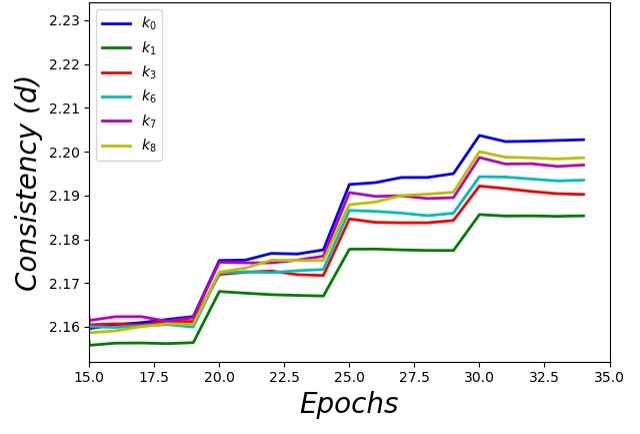}
        \label{fig:consistency_our}
     \end{subfigure}
     \begin{subfigure}[Comparison]{0.32\textwidth}
         \centering
        \includegraphics[width=0.75\textwidth]{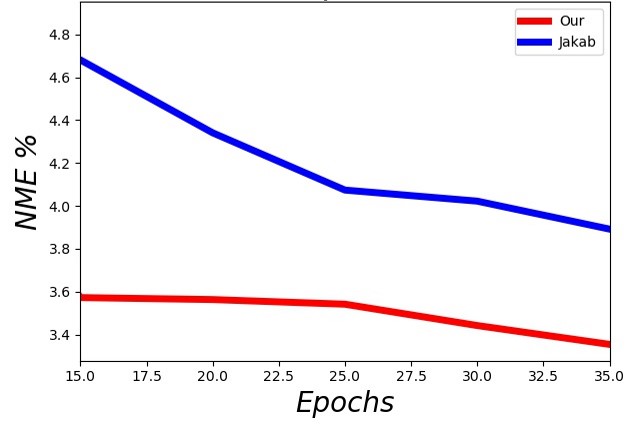}
        \label{fig:NME_comparison}
     \end{subfigure}
     \vspace{1em}
     \caption{Comparison of average landmark consistency via $d$. (a) Baseline (Jakab et al. ) (b) Ours  (c) the impact of $d$ on test forward error.}
    
    \label{fig: Consistency comparison} 
\end{figure*}

\begin{figure}[h!]
     \centering
 \includegraphics[width=.7\textwidth]{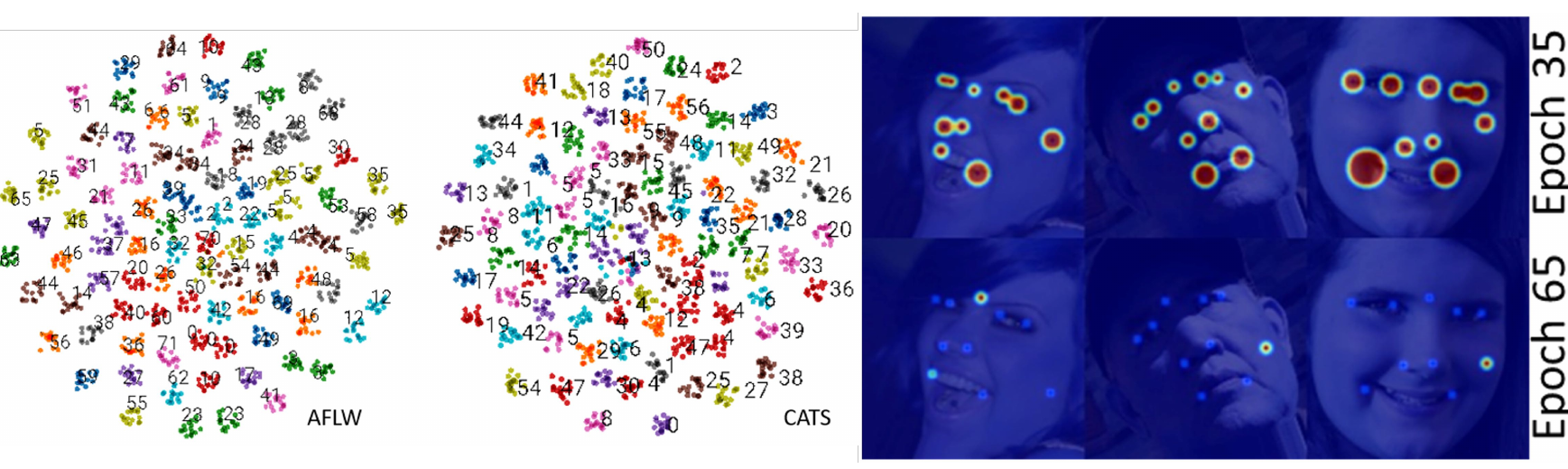}
         \label{fig:heatmaps}
\caption{Left: Clustered features using tSNE with cluster ids.  Right: Consistency-modulated heatmaps during training on AFLW. Larger blobs indicate lower consistency. }
    \label{fig:qual_all} 
\end{figure}

\noindent\textbf{Ablation Study and Analysis:} See suppl. for a study on method specific hyperparameters.\\
\noindent \textbf{On landmark consistency:} 
We compare landmark consistencies via the consistency measure $d$ during the training (Fig.~\ref{fig: Consistency comparison}). Our model learning strives to gradually improve landmark consistencies. In contrast, in baseline, the landmark consistencies remain almost the same during training. The landmark consistency also impacts (forward) error on test set and so in our case the improvement in landmark consistency is reflected by the decrease in the error. Fig.~\ref{fig:qual_all} (right) displays consistency-modulated heatmaps during training. Larger blob radius and higher redness indicate lower consistency. 

\begin{SCtable}[\sidecaptionrelwidth][h!]
    \centering
    \tabcolsep=0.05cm
    \small
    \scalebox{0.9}{
\begin{tabular}{lccp{0.1cm}ccp{0.1cm}ccp{0.1cm}cc}
\toprule
\multirow{3}{*}{Methods}  & \multicolumn{5}{c}{AFLW}             && \multicolumn{5}{c}{Cats Head}  \\ 
\cmidrule{2-6}\cmidrule{8-12}
&  \multicolumn{2}{c}{Epoch \# 65}                                          && \multicolumn{2}{c}{Epoch \# 110} & &  \multicolumn{2}{c}{Epoch \# 65}                                          && \multicolumn{2}{c}{Epoch \# 110}\\
\cmidrule{2-3}\cmidrule{5-6}\cmidrule{8-9}\cmidrule{11-12}
& Silh. &CH && Silh. &CH && Silh. &CH && Silh. &CH \\
\midrule
Kmeans           & -0.042        & 38.85         && -0.053       & 38.25          && 0.038         & 44.98          && -0.04         & 43.36          \\ 
KNN+GCN          & \textbf{0.723} & \textbf{296.1} && \textbf{0.74} & \textbf{337.4} &&\textbf{0.55} & \textbf{57.22} && \textbf{0.67} & \textbf{112.9} \\ 
\bottomrule
\end{tabular}
}
\caption{Quality of clustered landmark representations in our method  using Silhouette coefficient and Calinski-Harabasz (CH) Index.}
\label{tab:clustering_quality_comparisons} 
\end{SCtable}

\noindent \textbf{On landmark detector ($\Psi$) trained from scratch:} Tab.~\ref{tab:scratch_results} reports the performance of the baseline and our method when $\Psi$ is trained from scratch instead of being initialized from a checkpoint. Our method outperforms baseline by notable margins.

\noindent \textbf{Clustering Landmark Representations:} Fig.~\ref{fig:qual_all} (left) visualizes the clustered landmark features using t-SNE. 
The features are well-separated into different classes, and hence facilitate effective correspondence establishment. 
We also observe  clustering quality by KNN+GCN is much better than only Kmeans  (see Tab.~\ref{tab:clustering_quality_comparisons}). 
\begin{SCtable}[\sidecaptionrelwidth][h!]
    \centering
    \tabcolsep=0.05cm
    \small
    \scalebox{0.8}
    {
    \begin{tabular}{lccp{0.05cm}ccp{0.05cm}cc}
    \toprule
       \multirow{2}{*}{Methods/Datasets} 
       & \multicolumn{2}{c}{MAFL}  &&\multicolumn{2}{c}{AFLW}
       &&\multicolumn{2}{c}{Cats Head}\\
    \cmidrule{2-3}\cmidrule{5-6}\cmidrule{8-9}
    &Fwd & Bwd && Fwd & Bwd && Fwd & Bwd \\
    \midrule


        Baseline    & 6.27          & 16.6
                        &&  9.02        & 26.3
                        && 14.1         & 44.4 \\
        
                        
        Ours             & 3.92         & 8.49 
                        && 6.85        & 11.7 
                        && 4.1         & 3.41 \\

    \bottomrule
    \end{tabular} } 
 \caption{Performance of baseline and our method when the landmark detection network $\Psi$ is trained from scratch.}
    \label{tab:scratch_results} 
\end{SCtable}
\noindent \textbf{On pseudo-supervision:} Tab.~\ref{tab:kmeans_compare} evaluates the strength of our novel \textit{consistency-guided bottleneck formulation}, by replacing KNN affinity graph and refinement (KNN+GCN) with K-means for achieving pseudo-supervision. 


\begin{SCtable}[][!htp]
   \centering
   \tabcolsep=0.05cm
   \small
   \scalebox{0.7}{

   \centering
   \addtolength{\textfloatsep}{-0.2in}

   \begin{tabular}{lccp{0.05cm}ccp{0.05cm}cc}
   \toprule
      \multirow{2}{*}{Methods/Datasets} 
      & \multicolumn{2}{c}{MAFL}  &&\multicolumn{2}{c}{Cats Head}
      &&\multicolumn{2}{c}{LS3D}\\
   \cmidrule{2-3}\cmidrule{5-6}\cmidrule{8-9}
   &F & B && F & B && F & B \\
   \midrule

        Baseline\cite{jakab2018unsupervised}     & 3.99          & 4.53
                       && 4.53         & \textcolor{blue}{4.06}
                       && 5.38         & 7.06 \\
                        
       SOTA            & 3.99          & 4.53 
                       && 4.42         & 4.06 
                       && 5.38         & 6.57 \\
       Ours w/ KMeans         & \textcolor{blue}{3.73}          & \textcolor{red}{\textbf{3.90}}
                       && \textcolor{blue}{3.95}         & 4.95
                       && \textcolor{blue}{5.34}         & \textcolor{blue}{4.70} \\
       Ours w/ KNN+GCN        & \textcolor{red}{\textbf{3.50}}  & \textcolor{blue}{4.26}
                       && \textcolor{red}{\textbf{3.76}} & \textcolor{red}{\textbf{3.94}} 
                       && \textcolor{red}{\textbf{5.21}} & \textcolor{red}{\textbf{4.69}}  \\

   \bottomrule
   \end{tabular} } 
\caption{Comparison when either using KNN+GCN or K-means for pseudo-supervision with baseline \cite{jakab2018unsupervised} and SOTA methods \cite{jakab2018unsupervised,sanchez2019object,mallis2020unsupervised}. Red: Best, Blue: Second best.}
   \label{tab:kmeans_compare} 
   
\end{SCtable}

\begin{figure}[!htp]
  \centering
  
    \includegraphics[scale=0.2]{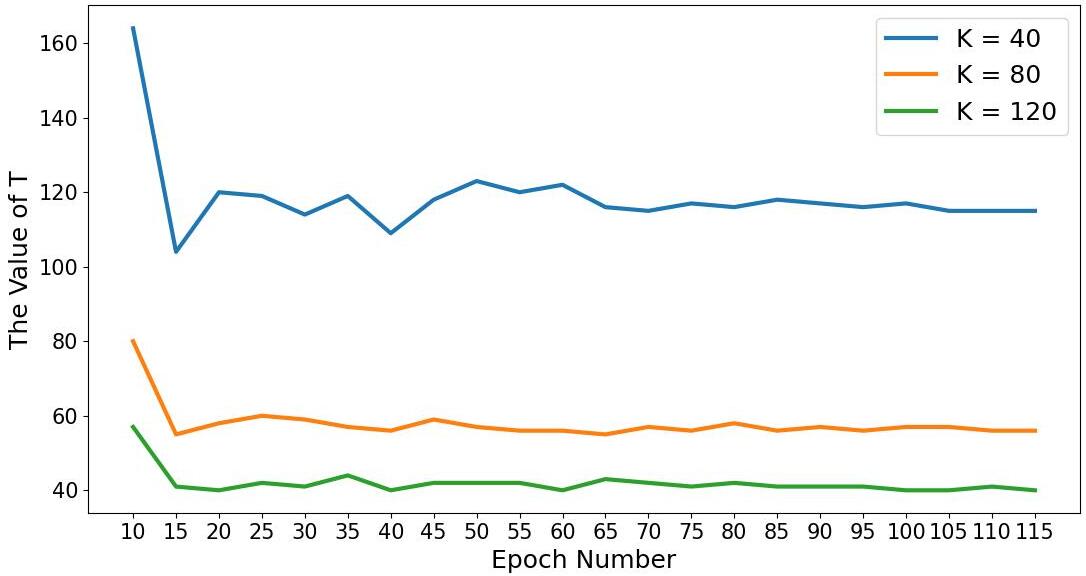}
   \caption{\small Evolution of $T$ for three different pre-fixed $\mathcal{K}$ values.} 
  
   \label{fig:T_evolution}
   \vspace{-3mm}
\end{figure}

We also plot the evolution of $T$ during training for different pre-fixed values of $\mathcal{K}$ in KNN (Fig.~\ref{fig:T_evolution}). We see that, for a given $\mathcal{K}$, the value of $T$ produced is less (approx by 20) than value of $\mathcal{K}$ throughout training.

\noindent Finally, we report both the forward and backward errors with different values of $\mathcal{K}$ (see Tab.~\ref{tab:k_value_comparison}). $\mathcal{K}$=80 used in our experiments, shows the best performance.

\begin{table}[!htb]
    \centering
    \scalebox{0.9}{
    \begin{tabular}{lccc}
    \toprule
         $\mathcal{K}$   & 40        & 80 (ours)          & 120 \\
         \midrule
         Forward (F) / Backward (B) & 6.29/6.71 & \textbf{5.91/6.39} & 6.20/7.23\\
         \bottomrule
    \end{tabular}}
    \vspace{0.5em}
    \caption{Performance with different values of $\mathcal{K}$.}
    \label{tab:k_value_comparison}
\end{table}


\section{Conclusion}
\label{sec:conclusion}
In this work, unsupervised landmark detection is improved by introducing a novel consistency-guided bottleneck. The landmark consistency is used for generating adaptive heatmaps. The consistency of a landmark is gauged by the proximity of its  representation to the cluster center considered as pseudo label.  Pseudo-supervision is established via landmark correspondence across multiple images. Extensive experiments on five publicly available datasets and a thorough analyses has demonstrated the effectiveness of the proposed approach. Excellent performance is observed compared to existing SOTA methods.

\bibliography{egbib}

\clearpage

\section{Supplementary}

\noindent \textbf{Landmark detector pre-training.}
For a fair comparison and following \cite{sanchez2019object}, the landmark detector $\Psi$ in our method, baseline \cite{jakab2018unsupervised}, and others with similar pipeline \cite{sanchez2019object} is initialised with the same checkpoint, pre-trained on MPII. Similarly, the VGG-16 network (in  the reconstructor) is pre-trained on ImageNet for our approach, baseline \cite{jakab2018unsupervised} and \cite{sanchez2019object}. 

\noindent \textbf{Image reconstruction network.}
For image reconstruction, we adapt from architectures typically used for image-to-image translation \cite{8100115}, face synthesis \cite{8578974,ma2018pose} and neural transfer \cite{johnson2016perceptual}.  We provide it with an image $\mathbf{y'}$ of resolution $128\times 128$, where $\mathbf{y'}$ is the deformed version of original image $\mathbf{y}$. We create this deformed image $\mathbf{y'}$ by applying random similarity transformations over image $\mathbf{y}$. These transformations include scaling, rotation and translation. We then proceed by first applying two downsampling convolutions that bring the number of features to 256, and then concatenate the adaptive heatmaps with the downsampled image tensor to pass it through a set of 6 residual blocks. Finally, we apply two spatial upsampling convolutions to restore the original image resolution.

\noindent \textbf{Evaluation metrics.}
We use \emph{forward} error \cite{sanchez2019object,mallis2020unsupervised}, \emph{backward} error \cite{sanchez2019object}, and Normalised Mean-squared Error (NME), normalized by inter-ocular distance to report the performance.
We train a linear regressor, that maps the discovered landmarks into the ground truth annotations, using a variable number of images in the training set. The learned regressor is tested on the corresponding test partition.
Following \cite{sanchez2019object,mallis2020unsupervised}, we refer to this as \emph{forward} error. In addition, \cite{sanchez2019object} also introduced a \emph{backward} error, that trains a regressor in an opposite direction. It maps the ground truth annotations into the discovered landmarks. We use Normalised Mean-squared Error (NME), normalized by inter-ocular distance to report the performance.

\begin{figure*}[t!]
    \centering    
    \includegraphics[width=0.97\textwidth]{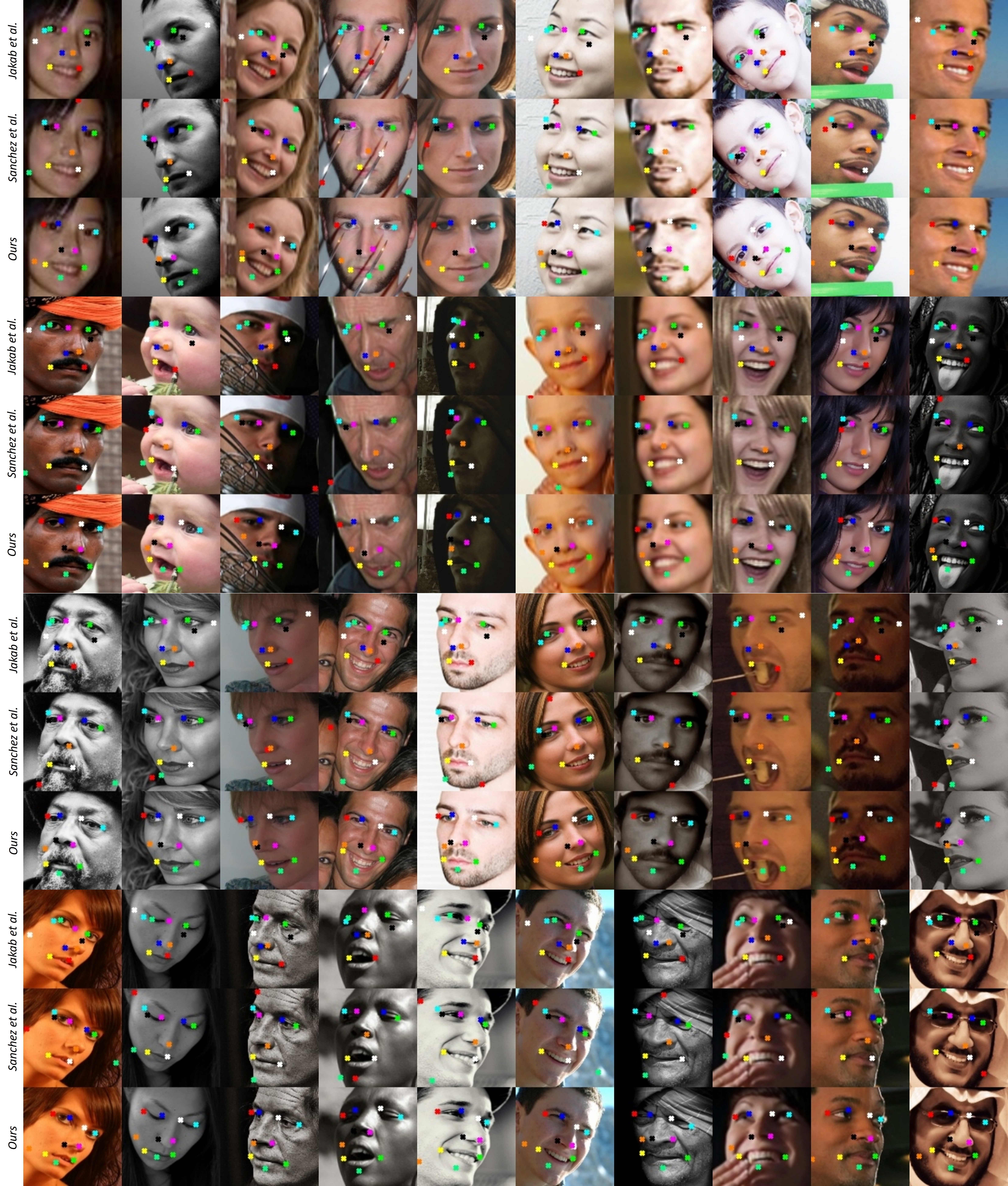}
    \caption{Additional qualitative comparisons on AFLW with Jakab et al. \cite{jakab2018unsupervised}(Baseline), and Sanchez et al. \cite{sanchez2019object}.
     }
    \label{fig:supp_aflw}
    \vspace{2ex}
\end{figure*}

\begin{figure*}[t!]
    \centering    
    \includegraphics[width=\textwidth]{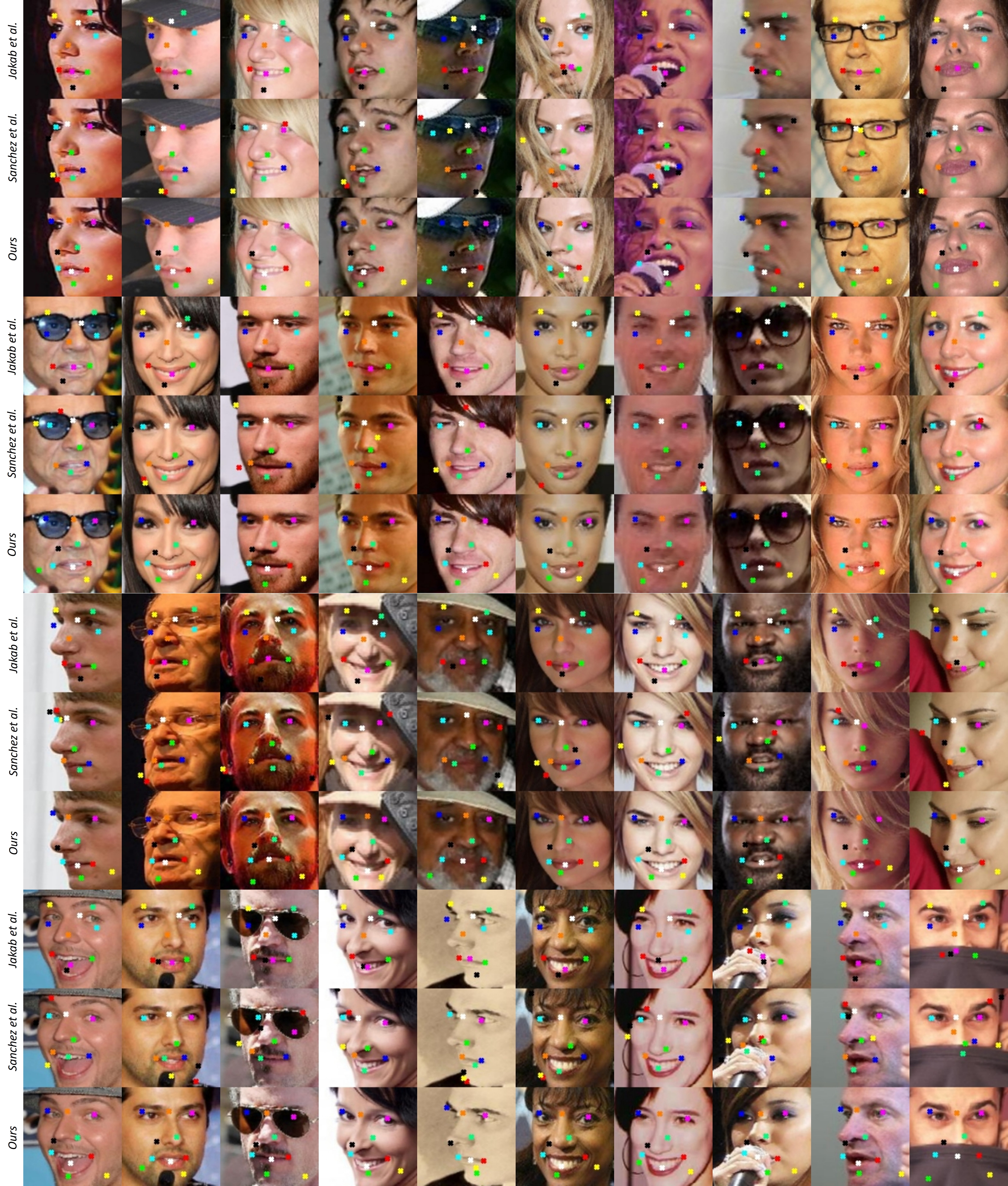}
    \caption{Additional qualitative comparisons on MAFL with Jakab et al. \cite{jakab2018unsupervised}(Baseline), and Sanchez et al. \cite{sanchez2019object}.
     }
    \label{fig:supp_mafl}
    \vspace{2ex}
\end{figure*}

\noindent\textbf{More qualitative results.}
Figs.~\ref{fig:supp_aflw} and~\ref{fig:supp_mafl} draw additional qualitative comparisons accompanying Sec.~4 (in main paper) on AFLW~\cite{koestinger2011annotated} and MAFL \cite{zhang2014facial} datasets. We see that our method is capable of discovering more semantically relevant landmarks that also capture improved correspondence across different poses and expressions. In contrast, other methods often detect semantically irrelevant landmarks that also lack appropriate correspondence across images. 

\begin{figure}[t!]
    \centering    
    \includegraphics[width=\textwidth]{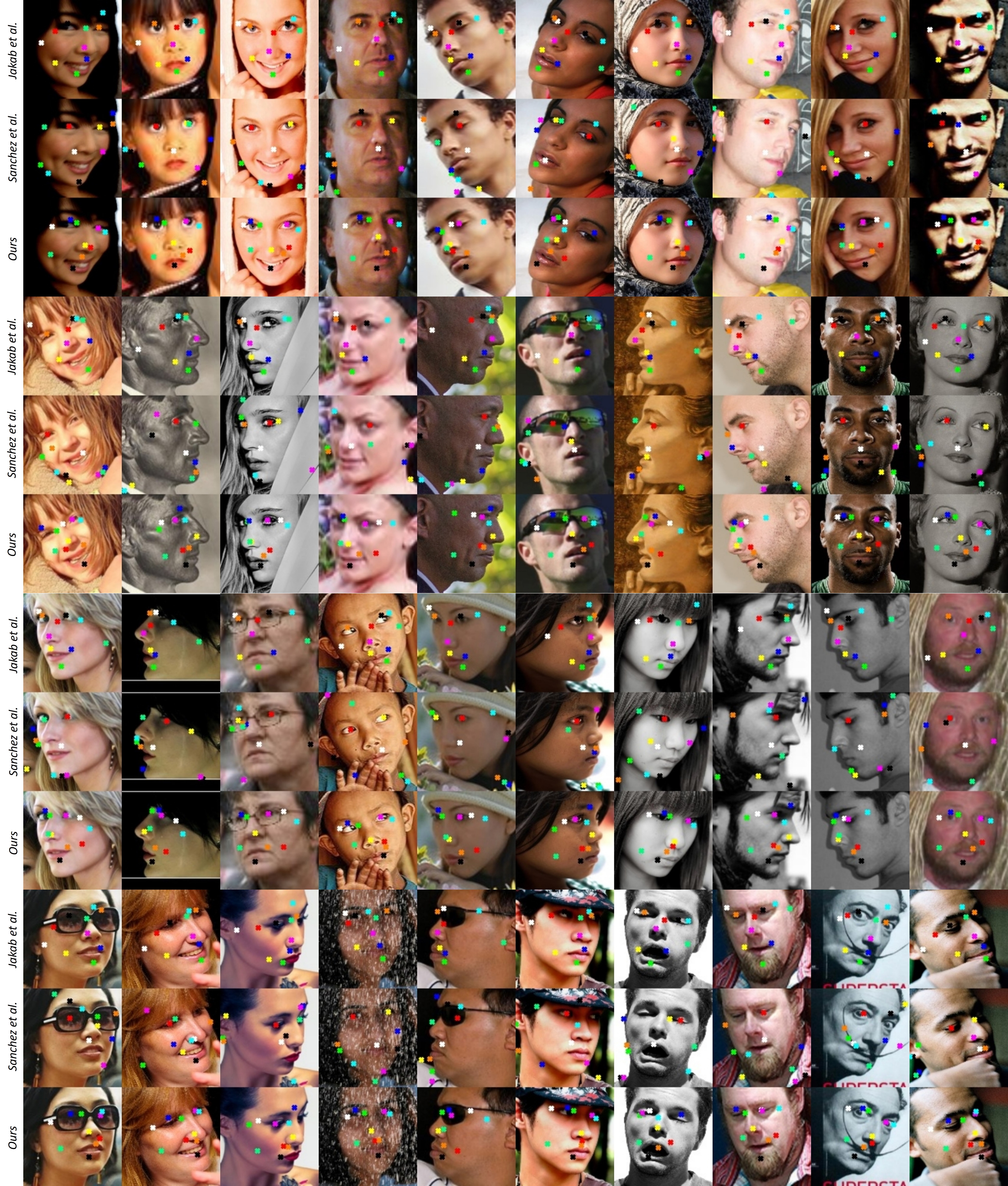}
    \caption{Additional qualitative comparisons on LS3D with Jakab et al. \cite{jakab2018unsupervised}(Baseline), and Sanchez et al. \cite{sanchez2019object}.
     }
    \label{fig:supp_ls3d}
    \vspace{2ex}
\end{figure}

\noindent Fig.~\ref{fig:supp_ls3d} shows qualitative comparisons in addition to Sec.~4 (in main paper) on LS3D \cite{bulat2017far} dataset. We can observe that, in contrast to other methods, our approach is able to discover more semantically meaningful under large pose and expression variations and other challenging factors such as occlusions.

\begin{figure}[t!]
    \centering    
    \includegraphics[width=\textwidth]{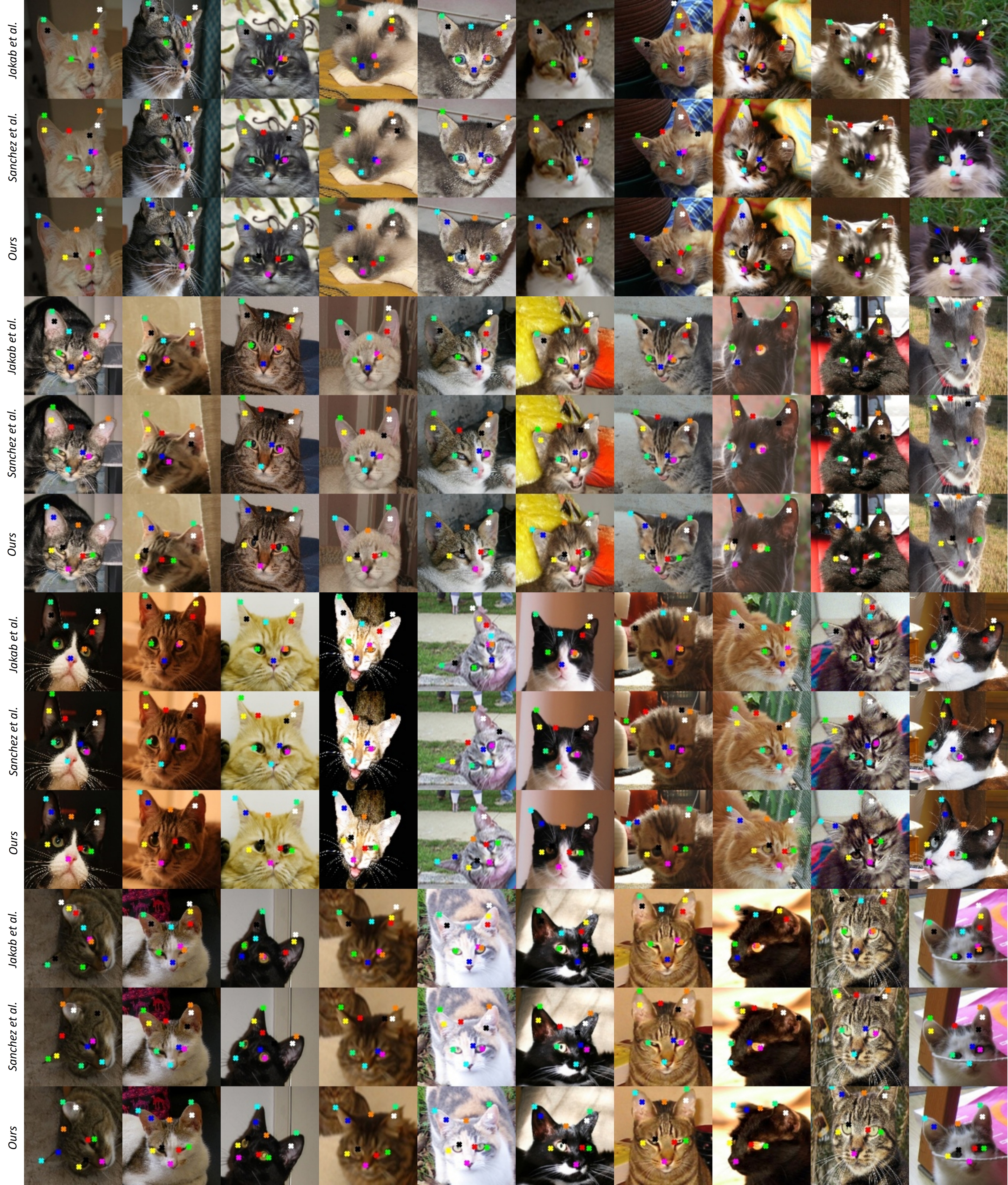}
    \caption{Additional qualitative comparisons on Cats Head with Jakab et al. \cite{jakab2018unsupervised}(Baseline), and Sanchez et al. \cite{sanchez2019object}.
     }
    \label{fig:supp_cats}
    \vspace{2ex}
\end{figure}

\begin{figure}[t!]
    \centering    
    \includegraphics[width=\textwidth]{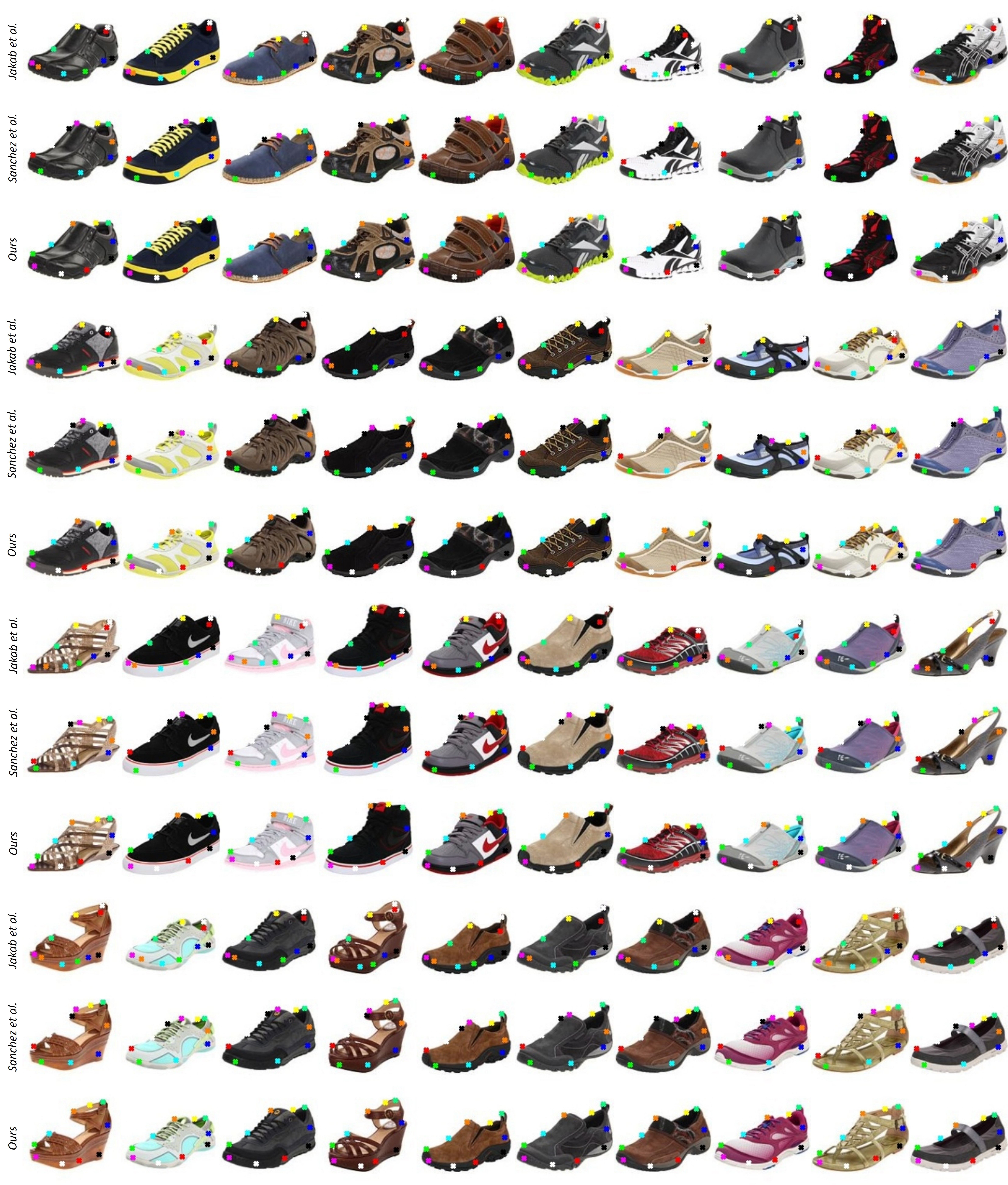}
    \caption{Additional qualitative comparisons on Shoes with Jakab et al. \cite{jakab2018unsupervised}(Baseline), and Sanchez et al. \cite{sanchez2019object}.
     }
    \label{fig:supp_shoes}
    \vspace{3ex}
\end{figure}

\noindent Figs.~\ref{fig:supp_cats} and~\ref{fig:supp_shoes} display additional qualitative comparisons on Cats Head \cite{zhang2008cat} and Shoes \cite{yu2014fine,yu2017semantic} datasets, respectively. In Cats Head dataset, in contrast to others, our method recovers semantically richer landmarks (e.g., around eyes and nose) under different appearance, pose and lighting variations.

\noindent\textbf{With another baseline.} We chose another competitive baseline using same loss function \cite{sanchez2019object} to evaluate the effectiveness of our proposed consistency-guided bottleneck (CGB). 
our CGB, also improves \cite{sanchez2019object} in both forward and backward errors (see Tab.~\ref{tab:sanchez_improvement_compare}).

\begin{figure}
\centering
\resizebox{0.7\linewidth}{!}{
\includegraphics[width=\linewidth]{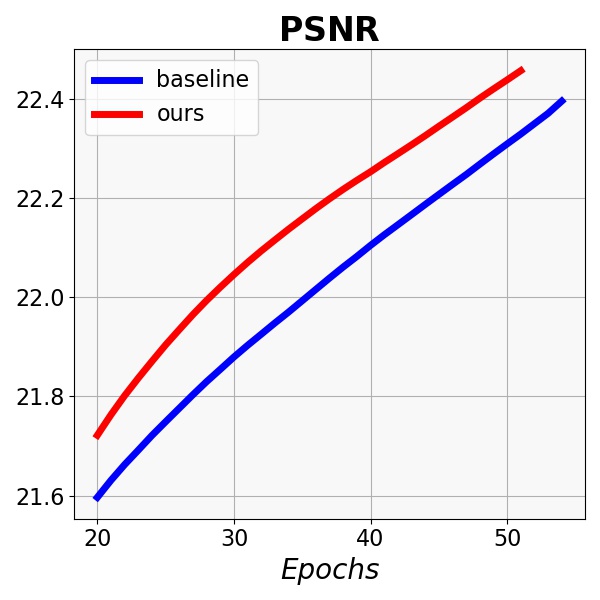}
\includegraphics[width=\linewidth]{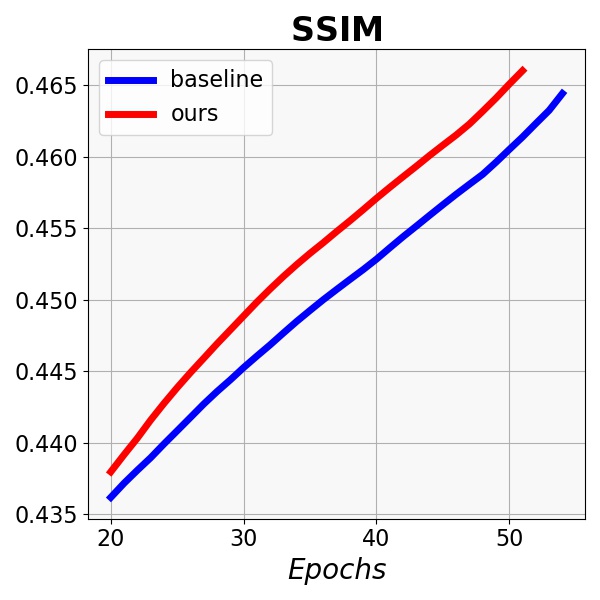}
}
\caption{\small Cumulative PNSR and SSIM \cite{wang2004image} over training (on Cats Head) to compare the reconstruction quality between our method and the baseline \cite{jakab2018unsupervised}.} 
    \label{fig:reconstruction_figures}
\end{figure}

\begin{table}[!htb]
\centering
 \begin{tabular}{lcccp{0.05cm}cccp{0.05cm}c}
 \hline
{Datasets} 
        & \multicolumn{2}{c}{AFLW}  &\multicolumn{2}{c}{MAFL}
        &\\  \hline
    Methods&F & B&F & B&\\
     \hline

        Sanchez\cite{sanchez2019object}     & 6.69          & 10.02
&3.99        &3.97\\

        Sanchez\cite{sanchez2019object}+Ours     & \textbf{6.29}          & \textbf{8.44} 
                       & \textbf{3.56}         & \textbf{3.76} \\
   \bottomrule
    \end{tabular}
\vspace{1em}
\caption{\small Our method is capable of boosting the performance of another competitive baseline \cite{sanchez2019object}.}
 \label{tab:sanchez_improvement_compare}
\end{table}
\noindent \textbf{Reconstruction quality comparison.} Fig.~\ref{fig:reconstruction_figures} shows that, compared to baseline \cite{jakab2018unsupervised}, our CGB allows improved reconstruction of the input image.

 \vspace{-0.2em}

\noindent \textbf{Varying the range of $\sigma$.} We study the impact on the performance upon varying the range of $\sigma$, Eq.(4) main paper, to which it is mapped (Table \ref{table:hyper parameter- range d}). Constraining the mapped range between $\lbrack0.2, 5\rbrack$ provides improved results compared to the relatively bigger range of $\lbrack0.2, 10\rbrack$. A much bigger range probably over dilates $\sigma$, which could likely degrade the reconstruction ability.

\noindent \textbf{Different manifestations of $\sigma$.} We report performance with different manifestations of $\sigma$: fixed, randomly sampled, and the modulated via landmark consistency (Table~\ref{tab:sigma_comparison}). Modulated $\sigma$ generally provides improved performance among others, thereby showing the effectiveness of favouring consistent landmarks over noisy counterparts during training.

\noindent \textbf{Varying pseudo-supervision update frequency.} We analyze performance upon varying the pseudo-supervision update frequency $\mathrm{PS_{update}}$ (Table \ref{table:hyper parameter GCN}). 

\noindent \textbf{Limitations.}
Like other SOTA methods (\cite{jakab2018unsupervised}, \cite{sanchez2019object}), our approach also depends on a pre-trained model trained in a supervised way on an object category. Further, the complexity of KNN affinity graph scales rapidly with more data points.
As such, this allows learning some structured representation, presumably shared across different object categories, and hence it could be beneficial for unsupervised landmark discovery task.

\begin{table}[]
\begin{minipage}{.3\textwidth}
    \centering
    \centering
\tabcolsep=0.07cm
    \centering
    \tabcolsep=0.05cm
    \small
    \begin{tabular}{l c}
    \toprule
         Method &NME\% \\
    \toprule
    Fixed $\sigma$\cite{jakab2018unsupervised} & 3.99\\
    Random $\sigma$ & 4.21\\
   $\sigma$(Ours) & \textbf{3.50}\\
      \bottomrule
    \end{tabular}
    \vspace{1em}
     {\caption{NME\% (forward) with different manifestations of $\sigma$.}
       \label{tab:sigma_comparison}}

\end{minipage}
\qquad \quad
\begin{minipage}{.2\textwidth}
    \centering
\tabcolsep=0.07cm
    \centering
    \tabcolsep=0.05cm
    \small
      \begin{tabular}{l c}
    \toprule
         $\sigma$ & NME\%  \\
    \toprule
    \lbrack0.2, 5\rbrack & \textbf{3.50} \\
    \lbrack0.2, 10\rbrack & 3.61 \\
    \bottomrule
    \end{tabular}
    \vspace{1em}
    {\caption{NME\% (forward) with varying the range of $\sigma$ mapping. }
      \label{table:hyper parameter- range d}}

\end{minipage}
\qquad \quad
\begin{minipage}{.3\textwidth}
    \centering
    \tabcolsep=0.05cm
    \small
 \begin{tabular}{l c}
    \toprule
         $\mathrm{PS_{update}}$  &NME\% \\
    \toprule
    5 & 3.50\\
    10 & 3.86\\
    20 &  3.70\\
    40 & \textbf{3.36} \\
    
    \bottomrule
    \end{tabular}
    \vspace{1em}
    {\caption{ $\mathrm{PS_{update}}$  variations.}
       \label{table:hyper parameter GCN}}

\end{minipage}
\end{table}
\begin{minipage}{.1\textwidth}
    
\end{minipage}

\end{document}